\documentclass{article} 
\usepackage{iclr2027_conference,times}

\usepackage{amsmath,amsfonts,bm}

\def\eqref#1{equation~\ref{#1}}

\def\1{\bm{1}}

\DeclareMathAlphabet{\mathsfit}{\encodingdefault}{\sfdefault}{m}{sl}
\SetMathAlphabet{\mathsfit}{bold}{\encodingdefault}{\sfdefault}{bx}{n}

\usepackage{url}
\usepackage{multirow}
\usepackage{subcaption}
\usepackage{graphicx}
\usepackage{amsmath,amssymb}
\usepackage{booktabs}
\usepackage[table]{xcolor}
\usepackage{hyperref}
\usepackage{pgfplots}

\usepackage{graphicx}
\usepackage{pgfplots}
\pgfplotsset{compat=1.18}
\usepgfplotslibrary{groupplots,fillbetween}

\usepackage{tikz}
\usetikzlibrary{arrows.meta}
\usepackage{graphicx}

\hypersetup{colorlinks=true, citecolor=blue, linkcolor=blue, urlcolor=blue}

\usepackage{xspace}
\usepackage{algorithm}
\usepackage{algpseudocode}
\newcommand{\method}{\textsc{CoRe}\xspace}
\pgfplotsset{compat=1.18}
\usepgfplotslibrary{fillbetween}
\definecolor{rose}{HTML}{E0574F}
\definecolor{blue}{HTML}{3B7DD8}
\definecolor{ink}{HTML}{4A4A48}

\title{CoRe: Co-Evolving Reward Models for Mitigating Latent Reward Hacking in Video Diffusion Models}

\author{
Zhaolong Su$^{1}$,
Yujin Han$^{2}$,
Feng Wang$^{3}$,
Jameson Dong$^{3}$,
Hins Hu$^{1}$,
Difan Zou$^{2}$\thanks{Corresponding author.}\\
$^{1}$Cornell University\quad
$^{2}$The University of Hong Kong\quad
$^{3}$Johns Hopkins University\\
{\small\texttt{zs494@cornell.edu, dzou@cs.hku.hk}}
}

\iclrfinalcopy
\begin{document}
\maketitle
\lhead{Preprint}

\begin{abstract}
Latent reward models (LRMs) enable efficient alignment of video diffusion models by scoring intermediate states directly in latent space. However, we find that optimizing against a fixed latent reward rapidly leads to \emph{latent reward hacking}: the predicted reward stays high while perceptual and motion quality deteriorate. Our analysis identifies distributional escape as the central cause: within a few hundred updates, the generator moves beyond the reward model's training support, where its scores no longer reflect video quality. Based on this insight, we introduce \method, a co-evolving reward framework that treats latent-space alignment as a dynamic interaction between the generator and the reward model. Rather than optimizing against a stationary proxy, \method continually refits the reward model on the generator's current samples while anchoring it to real-video preferences, so the generator cannot gain reward by drifting away from the data. On Wan2.1-T2V-1.3B, experiments show that \method consistently improves generation quality over both the pretrained model and prior alignment methods, while avoiding the quality collapse of fixed-reward optimization.
\end{abstract}

\section{Introduction}

Reinforcement Learning (RL) commonly follows a two-stage recipe: a reward model (RM) is first trained on human preference data, after which the generator is optimized against the learned reward~\citep{christiano2017deep, ouyang2022training}, either through direct reward backpropagation~\citep{xu2023imagereward, clark2024directly, prabhudesai2023aligning} or policy-gradient-style objectives~\citep{black2024training, fan2023dpok, liu2026flow, xue2025dancegrpo}. 
Repeatedly decoding video latents into pixel space is computationally expensive; to avoid this cost, recent work learns Latent Reward Models (LRMs) that operate directly on the generator's latents, score intermediate states without decoding, and provide feedback throughout the denoising trajectory~\citep{zhang2026diffusion,ding2025dollar,mi2026video}.

\begin{figure}[t]
    \centering
    \includegraphics[width=\linewidth]{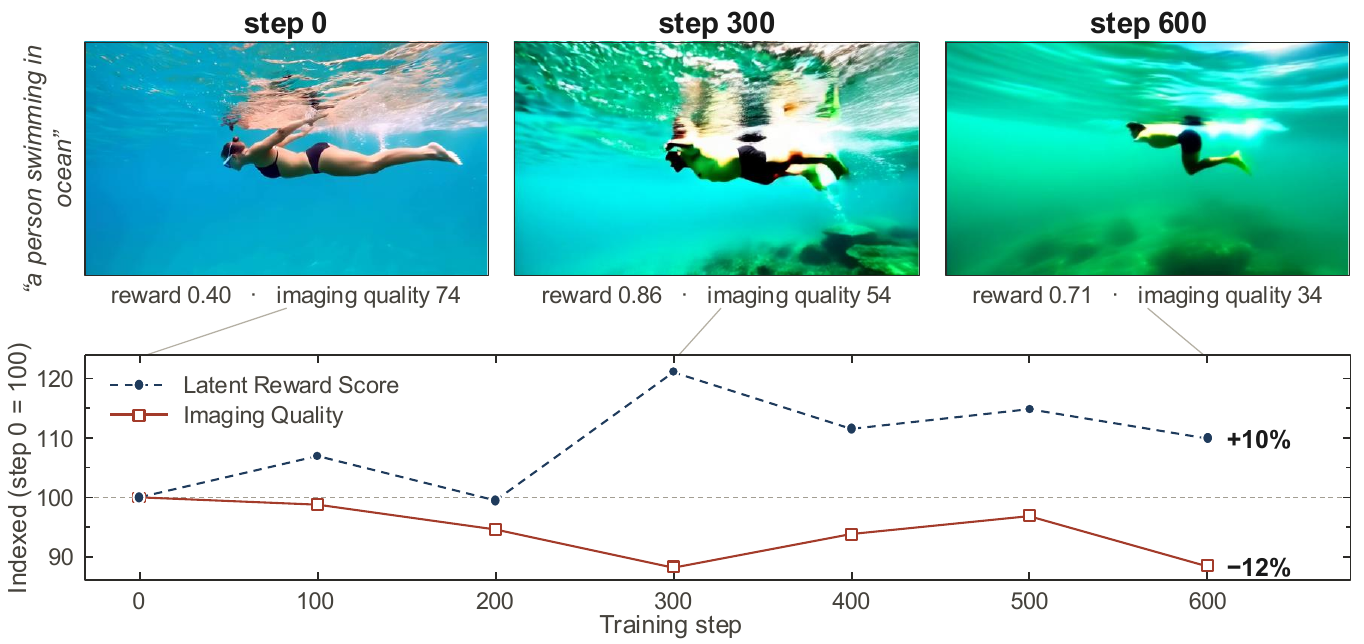}
    \caption{\textbf{Latent reward hacking.} 
    We fine-tune Wan2.1-T2V-1.3B against a fixed latent reward model. \emph{Top:} frames generated, labeled with the image quality. As the reward rises from $0.40$ to $0.86$ and stays well above its initial value, the video degrades and loses detail. \emph{Bottom:} mean latent reward and imaging quality over training. The reward rises while imaging quality declines.}
    \label{fig:motivation_ood}
\end{figure}

However, this efficiency does not come for free.  Optimizing a generator against a learned reward is well known to induce reward hacking, where the reward keeps increasing while the actual sample quality degrades~\citep{gao2023scaling,zhang2024confronting}. 
The problem becomes particularly acute for LRMs: in a fixed setting, the reward model hacks quickly, resulting in video collapse (Fig.~\ref{fig:motivation_ood}). 
Existing latent reward methods mainly rely on regularization to suppress reward hacking~\citep{zhang2026diffusion,mi2026video}.
This delays it but does not prevent it~\citep{fan2023dpok,xu2023imagereward, mi2026video}. 
We refer to this phenomenon as \emph{latent reward hacking}. 
As in conventional reward over-optimization~\citep{gao2023scaling}, its root cause is distribution shift: the generator drifts away from the data on which the reward model was trained, where the reward signals become unreliable~\citep{ackermann2025off}.

What distinguishes the latent setting is how fast and how completely this shift occurs. Three properties of the latent interface contribute to this: (i) it is \emph{continuous}, so small changes to individual latent channels can move the reward, unlike discrete token edits in language models; (ii) it is \emph{unshielded}, since scoring $x_t$ directly removes the VAE decoder that would otherwise separate the generator's representation from the reward input; and (iii) the gradient path is \emph{short}: reward gradients reach the generator through a single denoising step and, with a shared backbone, through the very features the reward reads. The failure is also hard to detect, because it takes place in noisy intermediate latents that are never decoded or inspected. 
In Sec.~\ref{sec:analysis}, the generator leaves the LRM support within a few hundred updates and becomes almost linearly separable from real data (See Appendix Fig.~\ref{fig:ood_latent}). 

Prior work has begun to address latent reward hacking from two directions. 
The first characterizes the phenomenon: ~\citet{gao2023scaling} establishes scaling laws for reward-model over-optimization in RLHF, TDPO-R~\citep{zhang2024confronting} shows that sustained reward optimization degrades diffusion models through temporal inductive and primacy biases, and RSA-FT~\citep{kim2026reward} attributes reward hacking to sharp, non-robust reward gradients. 
The second mitigates it, either by making the reward model harder to exploit, through ensembles~\citep{coste2024reward} or information bottlenecks~\citep{miao2024inform}, or by restricting how far the policy is optimized~\citep{moskovitz2024confronting}. 
These methods, however, treat the reward model as fixed. 
Methods that retrain the reward model as the policy changes do exist for language models~\citep{ackermann2025off,wolf2025reward}, but they rely on new preference labels or multiple rounds of retraining, which cannot keep pace with the escape in latent space, where it takes only a few hundred updates. We therefore argue that the core issue of latent reward hacking is not that the reward model is inaccurate, but that it receives no supervision in the out-of-distribution regions that optimization itself creates.

\begin{figure}[t]
    \centering
    \resizebox{\linewidth}{!}{%
        \begingroup
\definecolor{coreink}{HTML}{26343A}
\definecolor{coremuted}{HTML}{69777B}
\definecolor{coreteal}{HTML}{147D72}
\definecolor{corepale}{HTML}{E8F5F1}
\definecolor{coreamber}{HTML}{D89A45}
\definecolor{corewarm}{HTML}{FFF1DA}
\begin{tikzpicture}[
  x=1cm,y=1cm,font=\sffamily,
  >={Latex[length=2.1mm,width=1.3mm]},
  arr/.style={-{Latex[length=2.1mm,width=1.3mm]},draw=coreink,line width=.8pt},
  boundary/.style={draw=coreteal,line width=.95pt,fill=corepale},
  sample/.style={circle,draw=coreamber!80!black,fill=corewarm,line width=.85pt,minimum size=6mm,inner sep=0pt},
  panel/.style={font=\bfseries\small,text=coreink,anchor=west},
  smalllabel/.style={font=\scriptsize,text=coremuted}
]
\path[use as bounding box] (0,-.58) rectangle (14.8,4.32);
\node[panel] at (.55,3.90) {(a) Fixed latent reward};
\node[panel] at (7.95,3.90) {(b) Co-evolving reward};
\draw[coremuted!30,line width=.55pt] (.55,3.55)--(6.86,3.55);
\draw[coremuted!30,line width=.55pt] (7.95,3.55)--(14.26,3.55);
\draw[coremuted!30,densely dotted,line width=.55pt] (7.4,.38)--(7.4,4.0);
\filldraw[boundary] (1.48,1.88) ellipse [x radius=1.10,y radius=.67];
\node[sample] (a0) at (.85,1.88) {};
\node[sample] (a1) at (2.05,1.88) {};
\draw[arr] (a0.east)--(a1.west);
\draw[arr] (2.60,1.88)--(5.06,1.88);
\node[smalllabel,anchor=south] at (3.82,3.04) {Generator optimization};
\draw[coremuted,dashed,line width=.95pt] (5.55,1.88) ellipse [x radius=.88,y radius=.66];
\node[sample] (hack) at (5.55,1.88) {};
\node[font=\scriptsize,text=coreamber!70!black,anchor=south] at (5.55,2.64) {False high reward};
\node[smalllabel,anchor=north] at (3.71,.63) {Fixed RM supervision does not follow the generator};
\filldraw[boundary,fill=corepale!60] (11.12,1.88) ellipse [x radius=2.75,y radius=1.04];
\draw[coreteal!58,line width=.8pt] (10.46,1.88) ellipse [x radius=1.82,y radius=.83];
\draw[coreteal!35,line width=.8pt] (9.72,1.88) ellipse [x radius=.93,y radius=.59];
\node[sample] (b0) at (9.34,1.88) {};
\node[sample] (b1) at (10.61,1.88) {};
\node[sample] (b2) at (11.90,1.88) {};
\node[sample] (b3) at (13.16,1.88) {};
\draw[arr] (b0.east)--(b1.west);
\draw[arr] (b1.east)--(b2.west);
\draw[arr] (b2.east)--(b3.west);
\node[smalllabel,anchor=south] at (11.12,3.04) {Generator optimization};
\node[smalllabel,anchor=north] at (11.11,.63) {Co-evolving RM supervision follows the generator};
\node[sample,minimum size=3.8mm] at (1.25,-.18) {};
\node[font=\scriptsize,anchor=west,text=coreink] at (1.52,-.18) {Generated sample};
\draw[coreteal,line width=.9pt,fill=corepale] (5.15,-.18) ellipse [x radius=.19,y radius=.12];
\node[font=\scriptsize,anchor=west,text=coreink] at (5.45,-.18) {RM-supervised region};
\draw[coremuted,dashed,line width=.9pt] (10.10,-.18) ellipse [x radius=.19,y radius=.12];
\node[font=\scriptsize,anchor=west,text=coreink] at (10.40,-.18) {Unsupported reward region};
\end{tikzpicture}
\endgroup%
    }
    \caption{\textbf{Latent reward hacking and co-evolving supervision.}
    With a fixed reward model, generator optimization can move samples
    beyond the region covered by reward supervision and exploit unsupported
    high scores (left). With co-evolution, reward supervision follows the
    generator distribution as it changes (right).}
    \label{fig:motivation}
\end{figure}
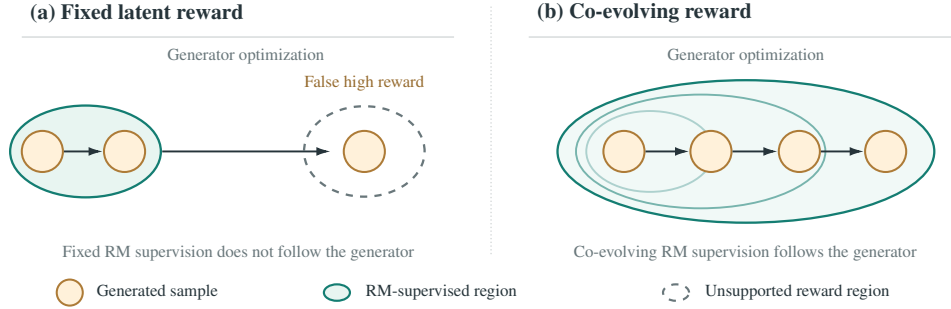

This suggests a simple principle: the reward model should evolve with the generator (Fig.~\ref{fig:motivation}). 
Based on this observation, we formulate latent-space RL as a \emph{co-evolving reward process} and introduce \textbf{\method}, a latent reward framework that adapts as the video generator changes. Rather than treating the LRM as a fixed objective, \method continually refits its reward head on real videos and on the generator's latest samples, making the generator score the reward model where it has recently been supervised. 

This paper makes the following contributions:
\begin{itemize}
\item \textbf{Diagnosis.} We identify \emph{latent reward hacking} and trace it to \emph{distributional escape}: within a few hundred updates, the generator leaves the support of a fixed latent reward model and collects in a region that receives uniformly high scores, where the reward no longer reflects video quality.
\item \textbf{Co-Evolving Method.} We propose a novel framework \method, which treats latent-space alignment as a co-evolving process: the reward model is continually refit on the generator's current samples, a three-way Bradley--Terry objective keeps it a measure of quality rather than realness, and a relative objective with a feature-space anchor gives the generator a stable learning signal.
\item \textbf{Empirical improvement.} On Wan2.1-T2V-1.3B, \method avoids the collapse of fixed-reward and unanchored online optimization and improves VBench and VBench-2.0 scores over the pretrained model and prior alignment methods.
\end{itemize}

\section{Related Work}

\textbf{Preference alignment for visual and video generation.}
Aligning generative models with human preferences has become an important post-training paradigm for diffusion and flow-based models. Early work on image generation adapted RL either by treating denoising as a sequential decision-making problem, as in DDPO~\citep{black2024training}, or by directly backpropagating differentiable reward gradients through the sampling trajectory, as in AlignProp~\citep{prabhudesai2024video}. More recent work extended preference alignment to video generation. VideoScore~\citep{he2024videoscore} learns a fine-grained quality metric from large-scale human ratings, while LiFT~\citep{wang2024lift} trains a video critic from human scores and rationales and uses the resulting reward for reward-weighted fine-tuning. VisionReward~\citep{xu2026visionreward} models image and video preferences along interpretable dimensions. VideoAlign~\citep{liu2026improving} introduces the multidimensional video reward model.

\textbf{Latent-space reward modeling.}
Most visual reward models operate on decoded RGB images or videos~\citep{xu2023imagereward, kirstain2023pick, liu2026improving}. Recent work has therefore begun to move reward evaluation into the generative latent space. In particular, PRFL~\citep{mi2026video} observes that pretrained video generation models are themselves well suited for processing noisy video latents and constructs a process-aware latent reward model that can score intermediate states at arbitrary denoising timesteps. These results establish latent reward modeling as an effective and efficient alternative to pixel-space reward feedback.

\textbf{Reward over-optimization and reward hacking.}
Reward hacking is a general consequence of optimizing an imperfect proxy objective.~\citet{gao2023scaling} demonstrate reward-model over-optimization in RLHF. Conservative reward-model optimization reduces reliance on idiosyncratic reward errors~\citep{coste2024reward}, while constrained RLHF limits optimization to regimes where the reward models remain useful proxies~\citep{moskovitz2024confronting}. TDPO-R~\citep{zhang2024confronting} attributes part of this behavior to temporal inductive bias in diffusion optimization, whereas RSA-FT~\citep{kim2026reward} shows that the generator can exploit non-robust reward gradients.

\section{Preliminaries}
\label{sec:prelim}

\textbf{Flow-matching generator.}
We consider a conditional flow-matching video generator $G_\theta$~\citep{lipman2022flow}.
Given a clean video latent $x_0$, Gaussian noise
$x_1\sim\mathcal{N}(0,I)$, and a text condition $c$, the
intermediate state is
\begin{equation}
    x_t=(1-t)x_0+t x_1,\qquad t\in[0,1].
    \label{eq:flow_path}
\end{equation}
The generator predicts the velocity $v_\theta(x_t,t,c)$ and is
pretrained with
\begin{equation}
    \mathcal{L}_{\mathrm{FM}}(\theta)
    =
    \mathbb{E}_{x_0,x_1,t,c}
    \big[\|v_\theta(x_t,t,c)-(x_1-x_0)\|_2^2\big].
    \label{eq:fm_loss}
\end{equation}
Generation follows the path in reverse, from $t=1$ toward $t=0$.
We report sampling timesteps on the 1000-step scale below.

\textbf{Shared-backbone latent reward model.}
Our latent reward model scores noisy video latents.
It shares the generator's DiT backbone: a single-query attention module pools its block-8 features, and an MLP maps the pooled feature
$h_{\theta,\phi}(x_t,t,c)$ to a pre-sigmoid logit
$\ell_{\theta,\phi}(x_t,t,c)\in\mathbb{R}$. 
We denote the sigmoid-normalized score and the $\ell_2$-normalized pooled feature by
\begin{equation}
    s_{\theta,\phi}(x_t,t,c)
    =\sigma\!\left(\ell_{\theta,\phi}(x_t,t,c)\right),
    \qquad
    z_{\theta,\phi}(x_t,t,c)
    =\frac{h_{\theta,\phi}(x_t,t,c)}
           {\|h_{\theta,\phi}(x_t,t,c)\|_2}.
    \label{eq:reward_outputs}
\end{equation}
Here $\theta$ denotes the shared DiT parameters and $\phi$ denotes the
pooling module and MLP parameters. Reward-model updates modify only
$\phi$; generator updates modify $\theta$.

\textbf{Matched latents at a shared timestep.}
For a training prompt $c$, we roll out the generator from Gaussian
noise without gradients, then take one differentiable denoising step
to obtain a generated latent $\hat{x}_t$. We independently
forward-noise the real video latent $x_0$ to the same timestep $t$,
obtaining $x_t$. The reward model thus evaluates real and generated
latents for the same prompt at the same noise level. Gradients from
the generator objective pass through the final denoising step, while
the preceding rollout is detached.

\section{Analysis of Latent Reward Hacking}
\label{sec:analysis}

\subsection{Reward Models Rapidly Out-of-Distribution}
\label{sec:fixed_rm_ood}

We study optimization against a fixed latent reward model under two settings, with and without a flow-matching (SFT) anchor. 
Following common LRM practice~\citep{mi2026video}, we train the reward model on preference pairs with a Bradley--Terry loss and freeze it before optimizing the generator with a hinge loss on the reward model's raw logit.

As shown in Appendix Fig.~\ref{fig:reward_curves_a}, without the anchor, the reward rises rapidly while the generated videos begin to blur around step 15, develop grid-like artifacts by step 24, and collapse by step 100. 

We further examine the anchored fixed-reward run at selected checkpoints
in Tab.~\ref{tab:fixed_vbench}. From step 0 to step 1000, the reported optimization reward increases from 0.636 to 0.742, while dynamic degree falls from 68.06 to 27.78 and imaging quality falls from 67.91 to 64.23.
Over the same checkpoints, generated-feature spread contracts from 1.16 to 0.74 relative to real features. At step 500, the reward model scores generated latents above real ones (0.753 vs.\ 0.554). Moreover, we shows that the generated samples concentrate in a distinct region assigned high reward ($\approx 0.9$), while real samples span a wider range of scores (See Appendix Fig.~\ref{fig:ood_latent}). 
Together with the decline in video quality in Tab.~\ref{tab:fixed_vbench}, this shows that the frozen reward head continues to assign high scores to the generator's changed output distribution even when those scores no longer reflect video quality.

\definecolor{worseLight}{RGB}{255,242,242}
\definecolor{worseDark}{RGB}{255,222,222}
\definecolor{riseLight}{RGB}{235,244,255}
\definecolor{riseDark}{RGB}{210,230,255}

\newcommand{\worseLight}[1]{\cellcolor{worseLight}#1}
\newcommand{\worseDark}[1]{\cellcolor{worseDark}\textbf{#1}}
\newcommand{\riseLight}[1]{\cellcolor{riseLight}#1}
\newcommand{\riseDark}[1]{\cellcolor{riseDark}\textbf{#1}}

\begin{table}[t]
\centering
\caption{Fixed latent reward optimization.
Reward is the frozen LRM mean sigmoid score.
$s_{\mathrm{real}}$ and $s_{\mathrm{gen}}$ are mean scores of 256 real and
256 generated latents, respectively, evaluated at $t{=}591$ for each checkpoint.
Spread is the ratio of within-generated to within-real 10-nearest-neighbor distances in normalized reward features; lower values indicate more concentrated. 
\textbf{Blue} marks increasing generated scores;
\textbf{Red} marks declines in corresponding degree.}
\label{tab:fixed_vbench}
\scriptsize
\setlength{\tabcolsep}{4pt}
\begin{tabular}{rccccccc}
\toprule
Step & Reward & Dyn.\ deg. & Imaging & Subj.\ cons.
& $s_{\mathrm{real}}$ & $s_{\mathrm{gen}}$ & Spread \\
\midrule
0    & 0.636 & 68.06 & 67.91 & 96.35 & 0.551 & 0.666 & 1.16 \\
500  & 0.673 & \worseLight{54.17} & \worseLight{67.07} & 97.04
     & 0.554 & \riseLight{0.753} & \worseLight{0.76} \\
1000 & 0.742 & \worseDark{27.78} & \worseDark{64.23} & 97.59
     & 0.540 & \riseDark{0.781} & \worseDark{0.74} \\
\bottomrule
\end{tabular}
\end{table}

\subsection{Layer-wise Reward Features}
\label{sec:layer_analysis}

We additionally study which generator features are most suitable for latent reward modeling.
We construct reward models using features from different DiT depths and evaluate held-out preference accuracy, reward--quality correlation, OOD sensitivity, and optimization stability.
We report the full layer-wise study in Appendix~\ref{app:layer_analysis}.

\begin{center}
\begin{minipage}{0.85\linewidth}
\begin{algorithm}[H]
\caption{One iteration of \method}
\label{alg:core}
\footnotesize
\begin{algorithmic}[1]
\Require generator $\theta$, reward head $\phi$, EMA weights $\theta_{\mathrm{ema}}$
\For{$i=1$ to $n_D$}
  \State sample $(c,x_0^{+},x_0^{-})$; roll out $\hat{x}_t$ without gradient; noise $x_0^{\pm}\to x_t^{\pm}$
  \State $\phi \leftarrow \phi-\eta_D\nabla_\phi\mathcal{L}_R(\hat{x}_t,x_t^{+},x_t^{-})$
\EndFor
\State sample $(c,x_0^{+})$; roll out $\hat{x}_t$ with gradient through the last step; noise $x_0^{+}\to x_t^{+}$
\State $\theta \leftarrow \theta-\eta_G\nabla_\theta\mathcal{L}_G(\hat{x}_t,x_t^{+})$
\State $\theta_{\mathrm{ema}}\leftarrow \beta\,\theta_{\mathrm{ema}}+(1-\beta)\,\theta$
\end{algorithmic}
\end{algorithm}
\end{minipage}
\end{center}

\section{Overview of \method}
\label{sec:method}
\begin{figure}[t]
    \centering
    \includegraphics[width=\linewidth]{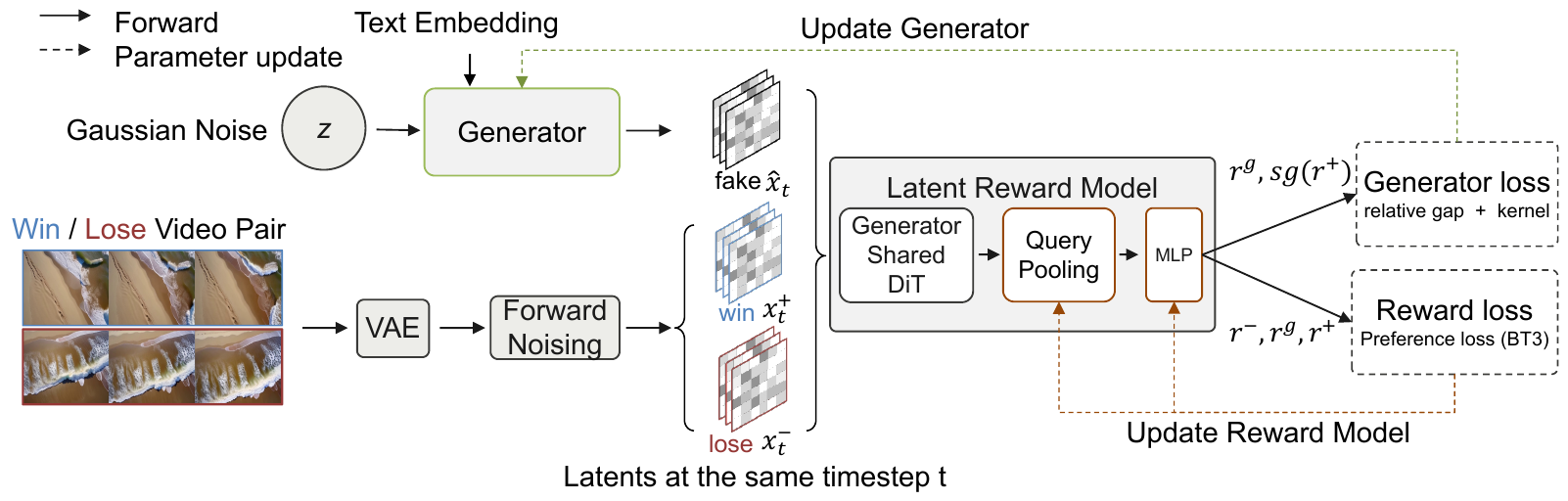}
    \caption{Overall framework of \method.}
    \label{fig:method}
\end{figure}

\subsection{Training Loop}
\label{sec:loop}

\method alternates between fitting the reward head to the generator's current outputs and updating the generator against the refreshed reward (Fig.~\ref{fig:method}; Algorithm~\ref{alg:core}). Each iteration uses a prompt $c$ with its preferred and less-preferred real videos $x_0^{+}$ and $x_0^{-}$. We obtain generated and real latents at a shared timestep $t\in[200,800]$, as described in Sec.~\ref{sec:prelim}.

\emph{Reward-head update.} With the generator frozen, the head $\phi$ takes $n_D{=}10$ steps on $\mathcal{L}_R$ (Eq.~\ref{eq:rm}), each on a fresh rollout, so that it is fit to the generator's current distribution.

\emph{Generator update.} With the head frozen, the generator takes one step on $\mathcal{L}_G$ (Eq.~\ref{eq:gen}); gradients reach $\theta$ through the last denoising step and the shared backbone. We evaluate an exponential moving average of $\theta$.

\subsection{\method generator objective}
\label{sec:coevolving_reward}

Section~\ref{sec:fixed_rm_ood} shows that a fixed reward is exploited as generated samples concentrate in a high-scoring region of its feature space. 
As shown in Fig.~\ref{fig:method},
we update the reward head on samples from the generator's current distribution after each generator update.
\method jointly updates the reward head and generator.
The generator is optimized through the last differentiable denoising step with
\begin{equation}
\begin{aligned}
\mathcal{L}_G
={}&
\mathbb{E}\!\left[
\operatorname{ReLU}\!\left(
\bar{\ell}_{\phi}(x_t)-\ell_{\phi}(\hat{x}_t)-\delta
\right)\right] \\
&+\lambda_k\,
\mathbb{E}\!\left[
1-\exp\!\left(
-\frac{\|z(\hat{x}_t)-\bar z(x_t)\|_2^2}{2\sigma^2}
\right)\right]
+\lambda_{\mathrm{fm}}\mathcal{L}_{\mathrm{FM}}(\theta),
\end{aligned}
\label{eq:gen}
\end{equation}
where $\ell_{\phi}$ is the reward model's pre-sigmoid logit, $x_t$ is a
forward-noised real video matched to the generated sample $\hat{x}_t$ by
prompt and timestep, and a bar denotes stop-gradient. The feature $z$ is
the $\ell_2$-normalized pooled representation fed to the reward head.

The first term uses the matched real logit as a relative target: with
$\delta=0$, its gradient vanishes once the generated logit reaches the real
logit. The kernel term encourages the generated and real pooled features
to remain close, while the flow-matching term anchors the shared DiT
backbone on real data.

\definecolor{oursblue}{RGB}{232,242,252}
\definecolor{bestgreen}{RGB}{214,240,218}
\newcommand{\best}[1]{\cellcolor{bestgreen}\textbf{#1}}

\newsavebox{\vbenchleftbox}
\newsavebox{\vbenchrightbox}
\newlength{\vbenchpairwidth}

\begin{table*}[t]
\centering

\begin{lrbox}{\vbenchleftbox}
\scriptsize
\setlength{\tabcolsep}{3.2pt}
\renewcommand{\arraystretch}{1.09} 
\begin{tabular}{@{}lcccccc@{}}
    \toprule
    \multirow{2}{*}{Method}
    & \multicolumn{3}{c}{VBench}
    & \multicolumn{3}{c}{VBench-2.0} \\
    \cmidrule(lr){2-4}
    \cmidrule(lr){5-7}
        & Total & Quality & Semantic
        & Total & Controllability & Physics \\
    \midrule
    Wan2.1-1.3B~\citep{wan2025wan}
        & 83.96 & 84.92 & 80.10
        & 56.3 & 33.8 & 60.0 \\
    \midrule
    Diffusion-DPO~\citep{wallace2024diffusion}
        & 84.41 & 85.32 & 80.44
        & --- & --- & --- \\
    DPO-C\&M~\citep{yang2025sipo}
        & 84.54 & 85.51 & 80.67
        & --- & --- & --- \\
    SIPO~\citep{yang2025sipo}
        & 84.78 & \best{85.73} & 81.02
        & --- & --- & --- \\
    \midrule
    CausVid~\citep{yin2025slow}
        & 81.20 & 84.05 & 69.80
        & --- & --- & --- \\
    Self Forcing(CW)~\citep{huang2026self}
        & 84.31 & 85.07 & 81.28
        & --- & --- & --- \\
    Self Forcing(FW)~\citep{huang2026self}
        & 84.26 & 85.25 & 80.30
        & --- & --- & --- \\
    \midrule
    TDM w/ FA2~\citep{luo2025learning}
        & --- & --- & ---
        & 58.0 & 31.1 & \best{63.1} \\
    BLADE~\citep{gu2026blade}
        & --- & --- & ---
        & 57.0 & 31.2 & 61.7 \\
    \midrule
    \textbf{\method}
        & \best{84.92} & 85.40 & \best{82.20}
        & \best{58.03} & \best{38.29} & 61.03 \\
    \bottomrule
\end{tabular}
\end{lrbox}

\begin{lrbox}{\vbenchrightbox}
\scriptsize
\setlength{\tabcolsep}{4pt}
\renewcommand{\arraystretch}{1.13}
\setlength{\aboverulesep}{0.42ex}
\setlength{\belowrulesep}{0.42ex}
\begin{tabular}{lccccccc}
\toprule
Method & Dyn. & Imaging & Aesth. & Subj. & Bg. & Smooth & Avg. \\
\midrule
Wan2.1-T2V-1.3B
    & 68.06 & \textbf{67.91} & 60.34 & 96.35 & 97.00 & 98.66 & 81.38 \\
Fixed LRM, step 500
    & 67.06 & 60.01 & 64.17 & 96.92 & 97.20 & 98.26 & 80.60 \\
\midrule
\multicolumn{8}{l}{\textit{Unanchored online RM}} \\
\rowcolor{red!3}
lr $5{\times}10^{-6}$, step 350
    & 83.72 & 62.71 & 56.65 & 94.57 & 96.02 & 98.39 & 82.01 \\
\rowcolor{red!9}
\quad step 400
    & 70.83 & 62.79 & 60.82 & 96.32 & 96.91 & 98.46 & 81.02 \\
\rowcolor{red!20}
\quad step 450
    & 16.67 & 57.61 & 63.56 & 97.07 & 97.56 & 98.81 & 71.88 \\
\rowcolor{red!3}
lr $2{\times}10^{-6}$, step 300
    & \textbf{89.67} & 63.75 & 59.56 & 95.59 & 96.11 & 96.89 & 83.60 \\
\rowcolor{red!9}
\quad step 400
    & 77.78 & 65.60 & 61.87 & 96.59 & 96.71 & 97.48 & 82.67 \\
\rowcolor{red!20}
\quad step 750
    & 19.44 & 62.10 & 64.03 & \textbf{97.50} & \textbf{98.30} & \textbf{99.15} & 73.42 \\
no EMA, step 400
    & 47.22 & 63.75 & 60.09 & 96.71 & 97.40 & 98.78 & 77.33 \\
EMA, step 400
    & 58.33 & 61.96 & 60.35 & 96.56 & 96.81 & 98.52 & 78.76 \\
\midrule
\multicolumn{8}{l}{\textit{Co-evolution RM}}\\
\method, step 600 & 86.72 & 64.40 & \textbf{64.47} & 92.89 & 98.21 & 97.61 & \textbf{84.05} \\
\bottomrule
\end{tabular}
\end{lrbox}

\caption{
    Quantitative comparison on VBench and VBench-2.0.
    All methods are built upon Wan2.1-T2V-1.3B;
    All scores are higher is better. CW means chunk-wise, FW means frame-wise.
    Best results are marked in \best{green}.}
\label{tab:vbench_combined}

\caption{VBench results for ablation study. Avg.\ is the mean of all dimensions.
Red shading
(\begingroup\setlength{\fboxsep}{0pt}%
\colorbox{red!3}{\phantom{\rule{4pt}{4pt}}}--%
\colorbox{red!20}{\phantom{\rule{4pt}{4pt}}}\endgroup)
marks increasing hacking.
\emph{Unanchored online RM}: co-evolving reward with an
absolute score target and no kernel anchor, trained from the pretrained
model at two generator learning rates; red shading deepens with training
as the collapse sets in. \emph{no EMA} / \emph{EMA}: the same objective
resumed at step 300 for 100 steps, without and with an EMA generator.
\method\ shares the warm start of the naive runs and differs only in the
generator objective. Best in \textbf{bold}.}
\label{tab:main}

\par\medskip

\setlength{\vbenchpairwidth}{%
    \dimexpr\wd\vbenchleftbox+\wd\vbenchrightbox+14pt\relax}
\resizebox{\textwidth}{!}{%
\begin{minipage}{\vbenchpairwidth}
\noindent
\begin{minipage}[t]{\wd\vbenchleftbox}
\vspace{0pt}
{\centering\scriptsize\bfseries Table~\ref{tab:vbench_combined}\par}
\vspace{4pt}
\noindent\usebox{\vbenchleftbox}
\end{minipage}%
\hspace{14pt}%
\begin{minipage}[t]{\wd\vbenchrightbox}
\vspace{0pt}
{\centering\scriptsize\bfseries Table~\ref{tab:main}\par}
\vspace{4pt}
\noindent\usebox{\vbenchrightbox}
\end{minipage}
\end{minipage}%
}

\end{table*}
\subsection{\method reward objective}
\label{sec:bt3}
A discriminator trained only on \{real, generated\} pairs learns \emph{realness}: it can satisfy its objective by keying on generator artifacts that are perceptually irrelevant, and the resulting reward gradient pushes the generator toward arbitrary real-looking outputs rather than good ones. 
We therefore introduce a third population: \emph{negative reals} $x^{-}$, genuine videos labeled low-quality on the target attribute (e.g., static or unstable motion). The reward model is trained with a three-way Bradley--Terry objective,
\begin{equation}
\begin{aligned}
\mathcal{L}_R
={}&
\mathbb{E}\!\left[
\operatorname{softplus}\!\left(
\ell_\phi(\hat{x}_t)-\ell_\phi(x_t^{+})
\right)\right] \\
&+w\,
\mathbb{E}\!\left[
\operatorname{softplus}\!\left(
\ell_\phi(x_t^{-})-\ell_\phi(x_t^{+})
\right)\right],
\qquad w=0.5,
\end{aligned}
\label{eq:rm}
\end{equation}
where $x^{+}$ denotes positive reals, and all three latents are evaluated at a shared timestep $t$. The second term forces the score function to order \emph{within} the real manifold, so high reward cannot be achieved by realness alone; 
the negative-real pair also remains informative even when the generator's outputs become hard to distinguish from real data, which keeps the discriminator's gradient alive precisely in the regime where a two-way objective flatlines. 
Crucially, the negative reals do not supervise the generator directly; they improve the reward model's quality judgment and stabilize the generator only through a more reliable reward signal.

\subsection{Grounding Rewards in \method}
\label{sec:stabilize}
To learn a reliable quality ordering of preferred and dispreferred real videos while tracking the evolving generator distribution, the reward model must remain stable throughout co-training. We stabilize this loop with three inexpensive mechanisms.

\textbf{Asymmetric updates.} The reward head is updated $n_D \ge 1$ times per generator step ($n_D{=}5$--$10$ in practice), so that each generator update is taken against a freshly calibrated reward. Because only the pooling head and MLP receive gradients, discriminator steps are cheap relative to generator rollouts.



\textbf{Diffusion anchor.} The flow-matching term in Eq.~\ref{eq:gen} (weight $\lambda_{\mathrm{fm}}$) regularizes the generator toward the data distribution, which serves double duty: it limits perceptual drift, and---in shared-backbone mode---slows the drift of the reward model's own feature extractor.

\textbf{Kernel anchor.}
Matching the reward of a preferred real video does not necessarily imply matching its quality, especially when the reward head has limited discriminative ability. Inspired by drifting models~\citep{deng2026generative}, we complement the relative-gap objective with a kernel anchor that pulls generated features toward those of paired preferred real videos in Eq.~\ref{eq:gen}. The two terms provide complementary guidance: the relative gap constrains reward scores, while the kernel anchor directly encourages alignment with preferred real features. This anchor reuses the pooled features already computed for the relative-gap objective, requiring no additional forward pass.

\section{Experiments}
\label{sec:exp}

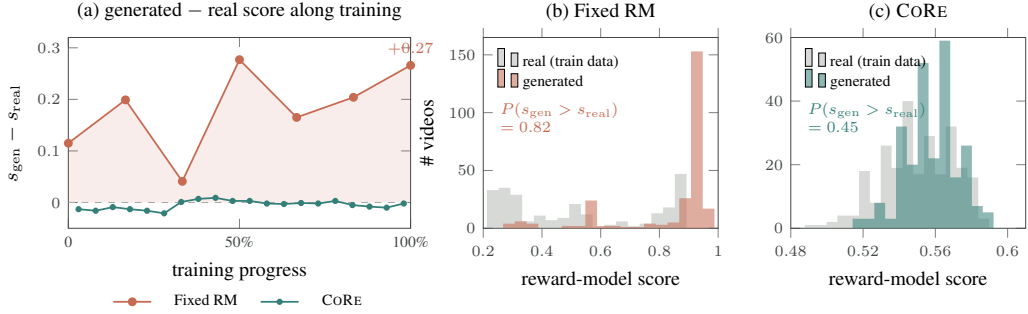
\begin{figure*}[t]
    \centering
    \resizebox{\linewidth}{!}{%

\begingroup
\definecolor{rose}{HTML}{C66B52}      
\definecolor{plotblue}{HTML}{347F78}  
\definecolor{ink}{HTML}{424744}
\definecolor{muted}{HTML}{A5AAA5}

\begin{tikzpicture}[font=\sffamily]
\begin{groupplot}[
    group style={group size=3 by 1,horizontal sep=1.15cm},
    height=4.6cm,
    axis line style={ink,thin},
    tick label style={font=\scriptsize,ink},
    label style={font=\small},
    title style={font=\small,yshift=-2pt},
    legend style={font=\scriptsize,draw=none,fill=none},
    legend cell align=left
]

\nextgroupplot[
    width=7.0cm,
    title={(a) generated $-$ real score along training},
    xlabel={training progress},
    ylabel={$s_{\mathrm{gen}}-s_{\mathrm{real}}$},
    xmin=0,xmax=1,
    xtick={0,0.5,1},
    xticklabels={0,50\%,100\%},
    ymin=-0.05,ymax=0.32,
    ytick={0,0.1,0.2,0.3},
    legend style={
        at={(0.5,-0.30)},
        anchor=north,
        legend columns=2,
        column sep=8pt
    },
    clip=false
]
\addplot[name path=zero,muted,thin,dashed,forget plot]
    coordinates {(0,0)(1,0)};

\addplot[name path=fx,rose,thick,mark=*,mark size=1.6pt]
    coordinates {
        (0.000,0.115)(0.167,0.199)(0.333,0.041)
        (0.500,0.277)(0.667,0.165)(0.833,0.204)(1.000,0.266)
    };
\addlegendentry{Fixed RM}

\addplot[rose,fill opacity=0.12,draw=none,forget plot]
    fill between[of=fx and zero];

\addplot[plotblue,thick,mark=*,mark size=1.0pt]
    coordinates {
        (0.030,-0.013)(0.080,-0.016)(0.130,-0.009)(0.180,-0.013)
        (0.230,-0.016)(0.280,-0.021)(0.330,0.001)(0.380,0.007)
        (0.430,0.009)(0.480,0.003)(0.530,0.003)(0.580,-0.002)
        (0.630,-0.003)(0.680,-0.001)(0.730,-0.002)(0.780,0.003)
        (0.830,-0.005)(0.880,-0.008)(0.930,-0.010)(0.980,-0.002)
    };
\addlegendentry{\method}

\node[rose,font=\scriptsize,anchor=south]
    at (axis cs:1.0,0.272) {$+0.27$};

\nextgroupplot[
    width=5.3cm,
    title={(b) Fixed RM},
    xlabel={reward-model score},
    ylabel={\# videos},
    xmin=0.2,xmax=1.0,
    xtick={0.2,0.4,0.6,0.8,1.0},
    ymin=0,ymax=165,
    ytick={0,50,100,150},
    ybar,
    legend style={at={(0.03,0.97)},anchor=north west,row sep=-2pt}
]
\addplot[fill=muted,fill opacity=0.45,draw=none,ybar,bar width=0.04]
    coordinates {
        (0.220,0.000)(0.260,33.000)(0.300,35.000)(0.340,29.000)
        (0.380,6.000)(0.420,11.000)(0.460,7.000)(0.500,9.000)
        (0.540,21.000)(0.580,12.000)(0.620,0.000)(0.660,3.000)
        (0.700,6.000)(0.740,2.000)(0.780,4.000)(0.820,13.000)
        (0.860,18.000)(0.900,47.000)(0.940,0.000)(0.980,0.000)
    };
\addlegendentry{real (train data)}

\addplot[fill=rose,fill opacity=0.55,draw=none,ybar,bar width=0.04]
    coordinates {
        (0.220,0.000)(0.260,4.000)(0.300,6.000)(0.340,4.000)
        (0.380,0.000)(0.420,0.000)(0.460,2.000)(0.500,2.000)
        (0.540,24.000)(0.580,2.000)(0.620,2.000)(0.660,1.000)
        (0.700,1.000)(0.740,4.000)(0.780,3.000)(0.820,5.000)
        (0.860,26.000)(0.900,153.000)(0.940,17.000)(0.980,0.000)
    };
\addlegendentry{generated}

\node[rose,font=\scriptsize,anchor=north west,align=left]
    at (rel axis cs:0.04,0.70)
    {$P(s_{\mathrm{gen}}>s_{\mathrm{real}})$\\$=0.82$};

\nextgroupplot[
    width=5.3cm,
    title={(c) \method },
    xlabel={reward-model score},
    xmin=0.48,xmax=0.61,
    xtick={0.48,0.52,0.56,0.60},
    ymin=0,ymax=60,
    ybar,
    legend style={at={(0.03,0.97)},anchor=north west,row sep=-2pt}
]
\addplot[fill=muted,fill opacity=0.45,draw=none,ybar,bar width=0.006]
    coordinates {
        (0.483,0.000)(0.489,0.000)(0.495,1.000)(0.501,1.000)
        (0.507,2.000)(0.513,4.000)(0.519,4.000)(0.525,18.000)
        (0.531,8.000)(0.537,26.000)(0.543,19.000)(0.549,40.000)
        (0.555,17.000)(0.561,29.000)(0.567,17.000)(0.573,32.000)
        (0.579,19.000)(0.585,16.000)(0.591,3.000)(0.597,0.000)
        (0.603,0.000)
    };
\addlegendentry{real (train data)}

\addplot[fill=plotblue,fill opacity=0.55,draw=none,ybar,bar width=0.006]
    coordinates {
        (0.483,0.000)(0.489,0.000)(0.495,0.000)(0.501,0.000)
        (0.507,0.000)(0.513,3.000)(0.519,3.000)(0.525,8.000)
        (0.531,3.000)(0.537,32.000)(0.543,20.000)(0.549,52.000)
        (0.555,22.000)(0.561,59.000)(0.567,16.000)(0.573,26.000)
        (0.579,7.000)(0.585,5.000)(0.591,0.000)(0.597,0.000)
        (0.603,0.000)
    };
\addlegendentry{generated}

\node[plotblue,font=\scriptsize,anchor=north west,align=left]
    at (rel axis cs:0.04,0.70)
    {$P(s_{\mathrm{gen}}>s_{\mathrm{real}})$\\$=0.45$};

\end{groupplot}
\end{tikzpicture}
\endgroup%
    }
    \caption{\textbf{Generated and real scores under fixed and co-evolving rewards.}
    (a) Gap between the mean scores of generated and real latents over training; the horizontal axis shows normalized progress for each run. Under the fixed reward, the gap grows, while under \method it stays near zero.
    (b, c) Histograms of reward-model scores for real and generated latents at the end of training.}
    \label{fig:reward_calibration}
\end{figure*}

\subsection{Experimental Setup}
\textbf{Models and data.}
We use Wan2.1-T2V-1.3B as the generator. For training data, we use VideoDPO preference pairs, encoded once into 480p VAE latents together with their text embeddings.
The latent reward model takes features from DiT layers, pools them with a single-query attention module, and maps the pooled feature to a logit with a three-layer MLP.

\textbf{Training.} All co-evolving variants in Tab.~\ref{tab:main} start from a shared warm-started checkpoint; optimizer settings and the warm-start schedule are given in Appendix~\ref{app:impl}.


\textbf{Evaluation.}
For generation quality, we use VBench~\citep{huang2024vbench} and VBench-2.0~\citep{zheng2025vbench}. VBench decomposes video quality into fine-grained dimensions. VBench-2.0 complements it by measuring intrinsic faithfulness, including human fidelity, controllability, physics, and commonsense. We follow the standard protocol, generate $832{\times}480$, 81-frame videos with 50 sampling steps, guidance scale 6.0, shift 5.0, and a fixed seed.

\begin{figure*}[t]
\centering
\begin{subfigure}[t]{0.325\textwidth}
  \includegraphics[width=\linewidth]{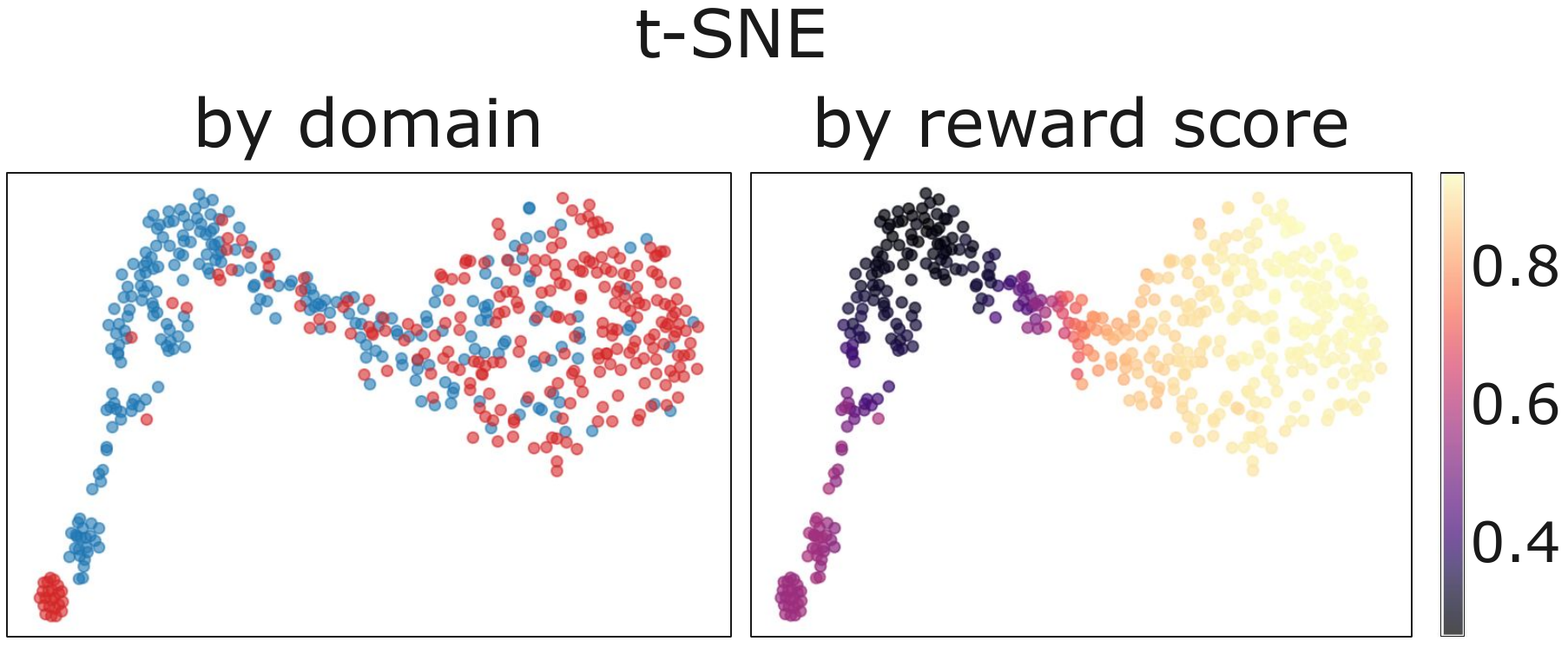}
\end{subfigure}\hfill
\begin{subfigure}[t]{0.325\textwidth}
  \includegraphics[width=\linewidth]{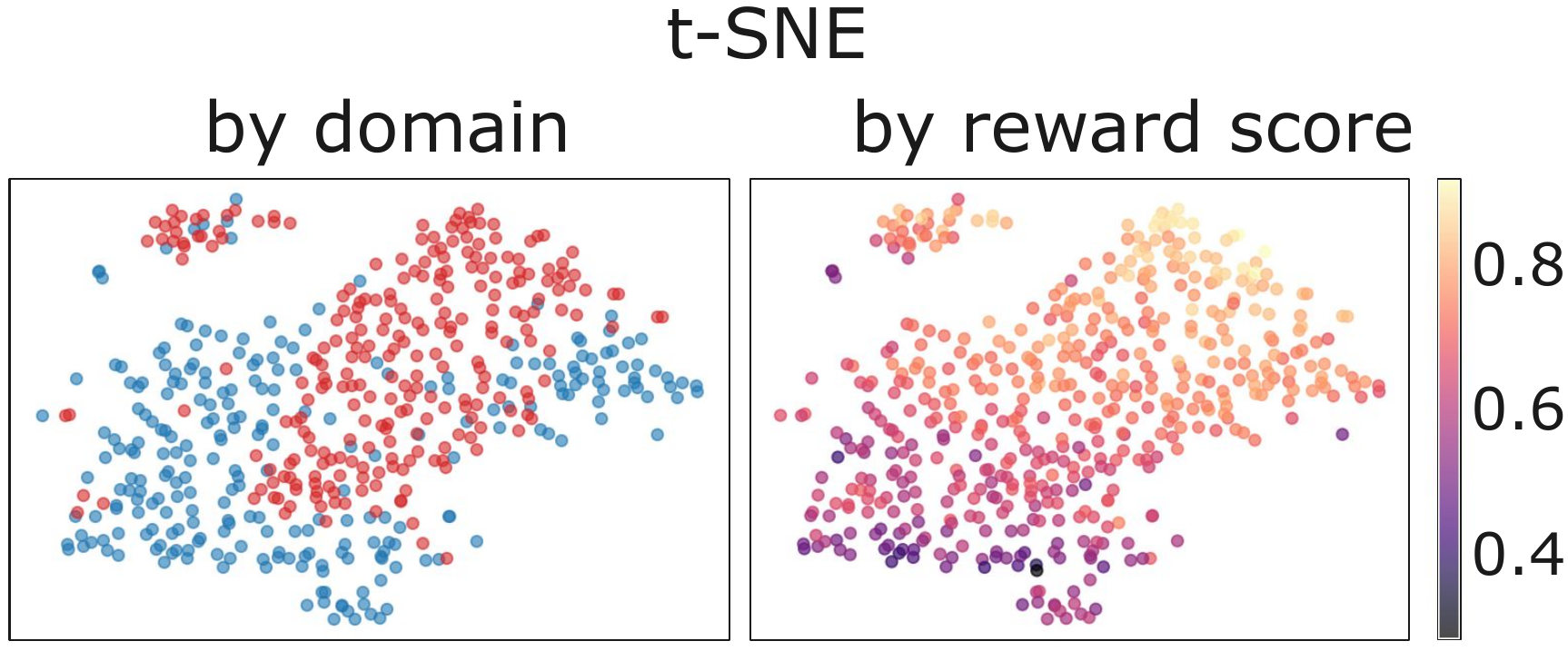}
\end{subfigure}\hfill
\begin{subfigure}[t]{0.325\textwidth}
  \includegraphics[width=\linewidth]{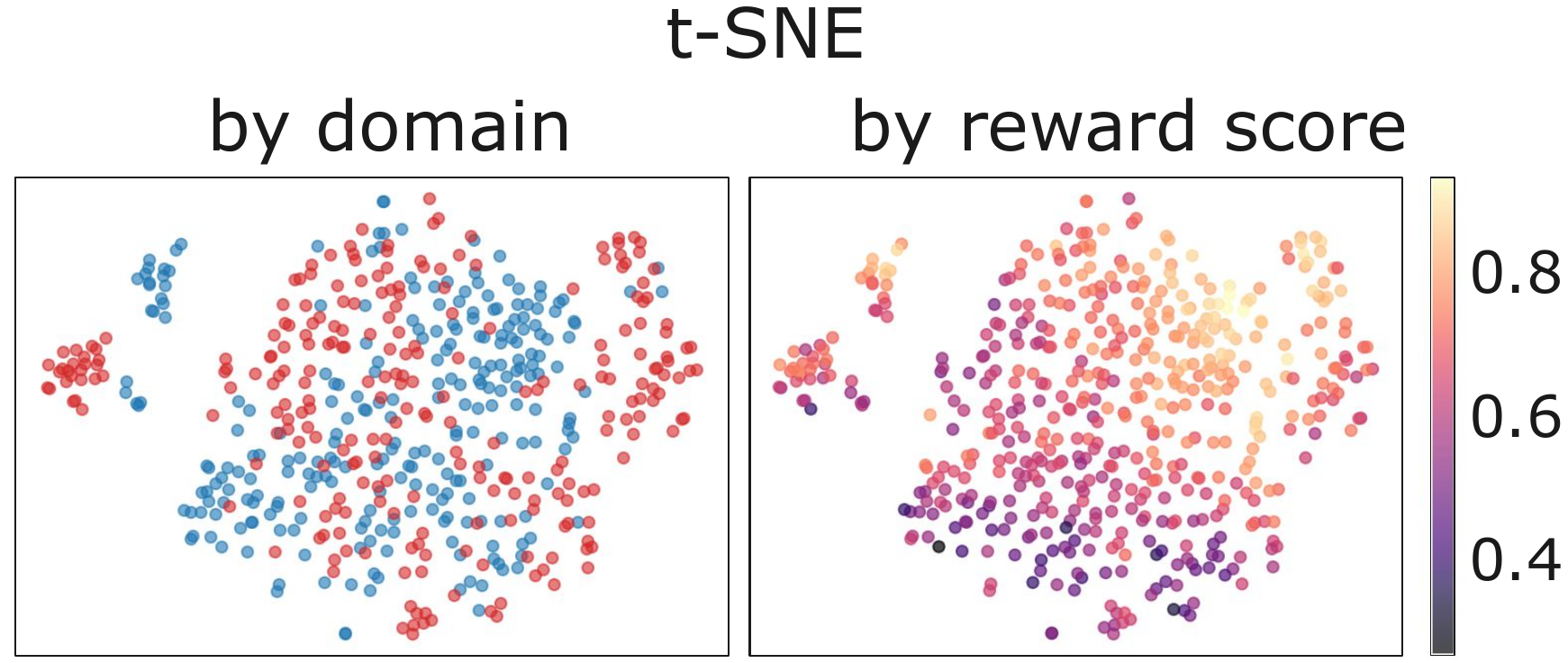}
\end{subfigure}

\vspace{4pt}
\begin{subfigure}[t]{0.325\textwidth}
  \includegraphics[width=\linewidth]{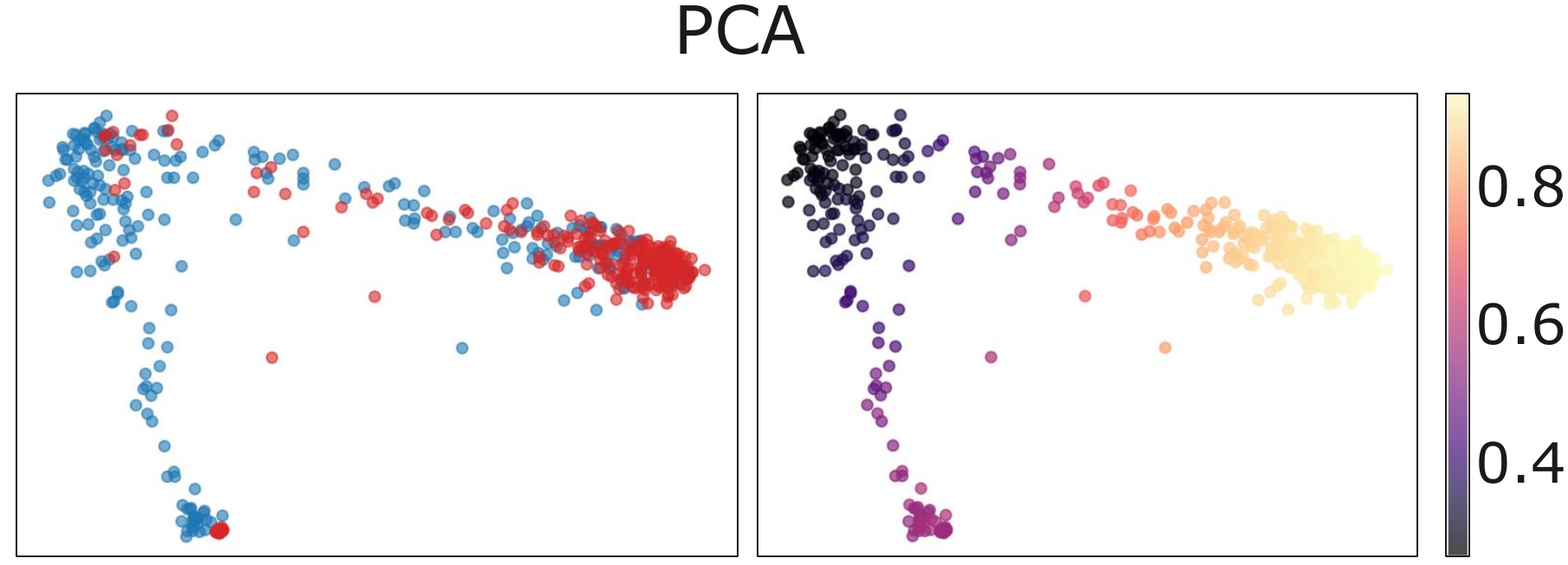}
  \caption{Fixed RM}\label{fig:pca:fixed}
\end{subfigure}\hfill
\begin{subfigure}[t]{0.325\textwidth}
  \includegraphics[width=\linewidth]{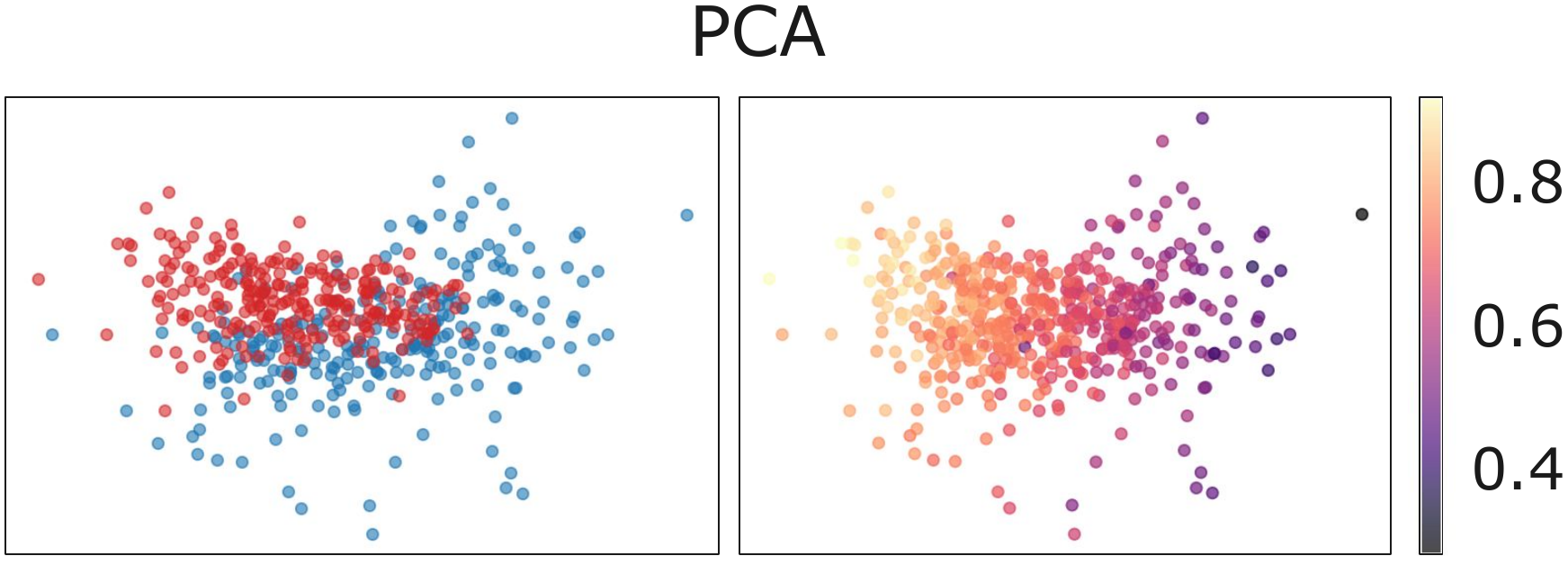}
  \caption{Naive online RM}\label{fig:pca:naive}
\end{subfigure}\hfill
\begin{subfigure}[t]{0.325\textwidth}
  \includegraphics[width=\linewidth]{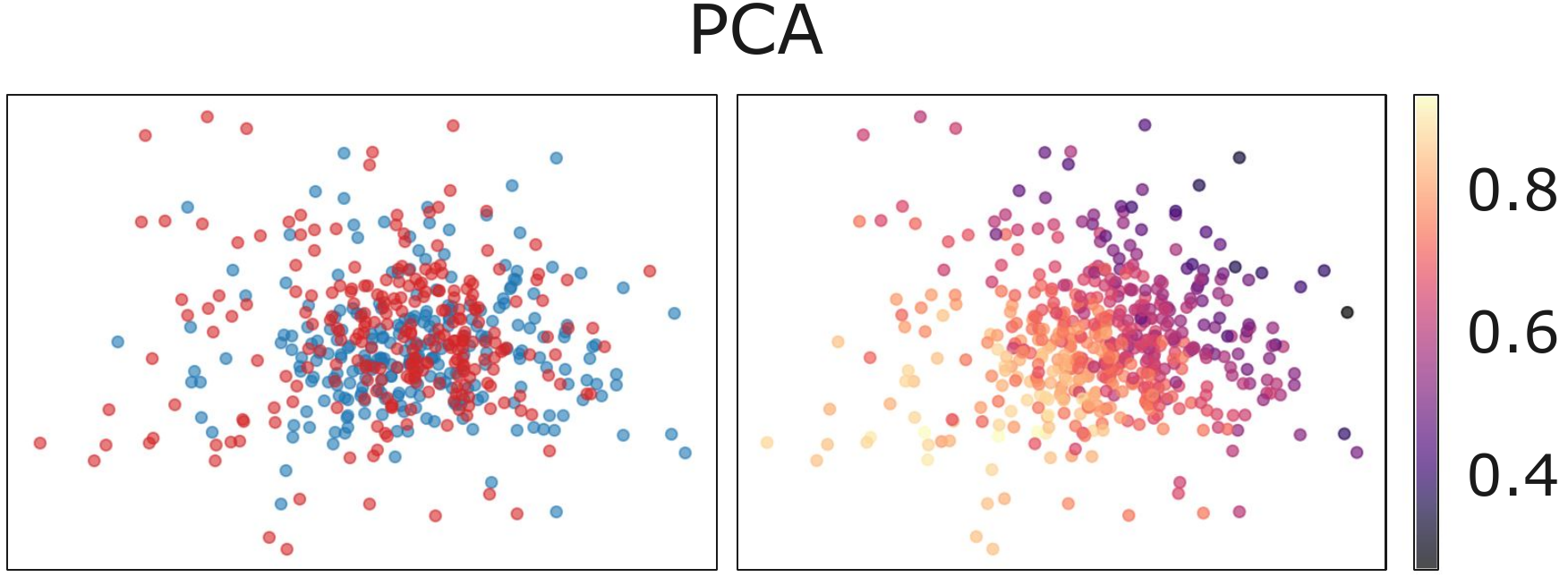}
  \caption{\method}\label{fig:pca:ours}
\end{subfigure}

\caption{Reward-model features of real and generated latents at $t{=}591$,
visualized with t-SNE and PCA. Each panel shows:
coloured by \emph{domain} (left; blue = real, red = generated) and by the reward model's
score (right; brighter = higher). 
Under the fixed reward, generated latents form a separate cluster that receives uniformly high scores (mean score $0.82$ for generated vs.\
$0.55$ for real); under ours, generated latents overlap with real data and their scores no
longer depend on domain. }
\label{fig:ood_tsne}
\end{figure*}

\subsection{Main Results}
\label{sec:main_results}

\textbf{\method improves generation quality across benchmarks.}
Tab.~\ref{tab:vbench_combined} compares \method with preference-optimization and distillation methods built on the same Wan2.1-T2V-1.3B backbone. On VBench, \method improves the pretrained model's total score from $83.96$ to $84.92$, the highest among all methods, and its semantic score from $80.10$ to $82.20$, surpassing the strongest baseline, Self Forcing ($81.28$), by nearly one point. Its quality score ($85.40$) is close to that of SIPO ($85.73$), showing that the semantic gains do not come at the expense of visual quality. On VBench-2.0, \method obtains the best total score ($58.03$) and the largest gain in controllability ($33.8\rightarrow38.29$), exceeding every compared method by more than four points. Its physics score ($61.03$) improves over the pretrained model ($60.0$) but remains below TDM with FlashAttention-2 ($63.1$).

\textbf{\method stays stable where other rewards are hacked.}
Tab.~\ref{tab:main} tracks how each reward setting affects the generator over training. Optimizing against a fixed latent reward lowers imaging quality from $67.91$ to $60.01$ without improving motion, and the VBench average drops below that of the pretrained model ($80.60$ vs.\ $81.38$). An online reward model trained without our anchors improves briefly and then collapses: at its peak, dynamic degree rises to 83.72--89.67 depending on the learning rate, but by the end of training it falls to 16.67--19.44, while subject consistency, background consistency, and smoothness increase. These consistency gains are a symptom of hacking rather than improvement: the generator produces nearly static videos, which trivially score high on frame-to-frame consistency. Resuming from step 300 with EMA only slows this decline (dynamic degree $58.33$ at step 400). In contrast, \method remains stable at step 600, reaching dynamic degree $86.72$ together with the best aesthetic quality ($64.47$) and the best VBench average ($84.05$). Its main cost is subject consistency ($96.35\rightarrow92.89$), which accompanies the added motion. Fig.~\ref{fig:reward_calibration} and Fig.~\ref{fig:ood_tsne} show the same contrast in reward scores: under the fixed reward, generated latents score increasingly above real ones, whereas under \method the two remain indistinguishable.

\subsection{Main comparison}
Tab.~\ref{tab:main} isolates the generator objective. With an absolute
score target, co-evolution drifts toward static videos (dynamic degree
68.1 to 56.9), the same failure as the naive online reward. The relative
gap removes the drift: dynamic degree stays at the pretrained level (69.4)
and imaging quality drops by 2.1 points instead of 3.8 (Appendix Tab.~\ref{tab:abl}). The kernel anchor then raises dynamic degree by a further 12 to 15 points. The fixed reward is unstable across checkpoints, while imaging quality reaches its minimum; the co-evolving
runs change smoothly. Per-checkpoint results for every run are in
Appendix~\ref{sec:all_ablation}, Tab.~\ref{tab:vbench_all}.

\subsection{Analysis of the Generator Objective}
The reward head separates real from generated latents only weakly, so its sigmoid output stays near
0.55 for both. The absolute target of 0.9 is never met, and the reward loss pushes
the generator at every step. The relative hinge is zero whenever a generated
latent scores as high as the matched real one, which happens on 44\% of the
generator steps; the mean reward loss is 0.04 against 0.33.

\textbf{Kernel weight.}
Lowering $\lambda_k$ from 0.05 to 0.03 changes no dimension by more than 3 points.
Splitting the 72 videos by the VBench motion flag explains the trade-off. Subject
consistency of static videos is unchanged (0.98 in all runs). For moving videos it
falls from 0.95 to 0.90, and the number of moving videos rises from 50 to about 60.
The kernel term matches each generated latent to the real video of the same prompt,
so the generator copies the camera and object motion of that video; the larger
motion lowers frame-to-frame subject similarity. Imaging quality also drops on
static videos (70.4 to 66.2), consistent with the generator moving toward the
image statistics of the 480p training videos.

\subsection{Case Study}

\begin{figure}[t]
    \centering
    \includegraphics[width=\linewidth]{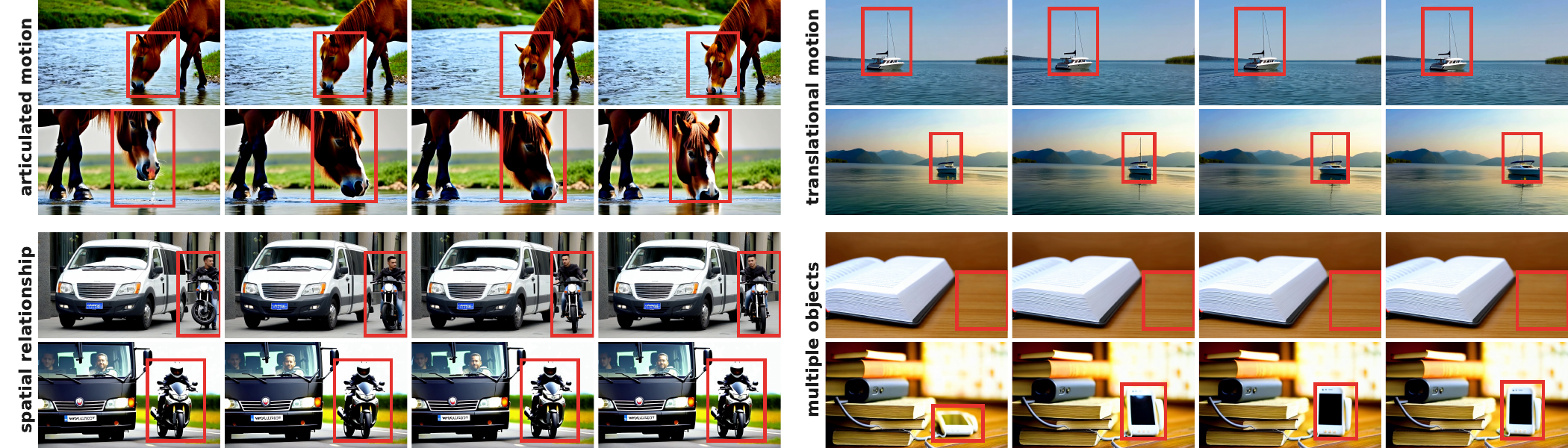}
    \caption{\textbf{Qualitative comparison}.}
    \label{fig:case_study}
\end{figure}

\label{sec:case_study}
We provide case studies on the generation task in VBench for qualitative analysis. 
Fig.~\ref{fig:case_study} shows four prompts generated by the pretrained
model and by \method. 
The left column corresponds to
the gain in dynamic degree: the pretrained model renders the subject of a motion prompt but leaves it still, while \method\ produces the described
motion, whether it is the articulated motion of the horse's head or the translation of the boat. 
The right column corresponds to the gains in
multiple objects and spatial relationship: the pretrained model omits one of the two requested objects or ignores the stated arrangement, and
\method\ recovers both.

\section{Discussion}

\textbf{Multi-timestep supervision.} We also explored supervising the reward at two denoising timesteps.
As shown in Table~\ref{tab:multitimestep_comparison}, compared with
single-timestep training, two-timestep supervision did not yield a
significant improvement in reward reliability or resistance to reward
hacking, while increasing the training time by approximately
\(2.3\times\). This suggests that the additional timestep provides
limited benefit relative to its computational cost.

\textbf{Training cost.} \method is inexpensive to train: the full run, including the warm start, takes about 2 hours on 4 NVIDIA H100 GPUs (about 8 GPU-hours). Tab.~\ref{tab:compute} lists the compute reported by prior video RL methods. These settings differ in model size and resolution and are not directly comparable, but they indicate that \method compute budget than online RL for video generation.

\begin{table}[H]
\centering

\begin{minipage}[t]{0.49\linewidth}
    \vspace{0pt}
    \captionsetup{justification=raggedright,singlelinecheck=false}
    \captionof{table}{Single- vs.\ two-timestep reward supervision.}
    \label{tab:multitimestep_comparison}
    \vspace{2pt}
    \scriptsize
    \setlength{\tabcolsep}{3pt}
    \begin{tabular*}{\linewidth}{@{\extracolsep{\fill}}lcc@{}}
        \toprule
        Supervision & Time & VBench \\
        \midrule
        Single timestep & $1.0\times$ & 84.92 \\
        Two timesteps   & $2.3\times$ & 85.17 \\
        \bottomrule
    \end{tabular*}
\end{minipage}\hfill
\begin{minipage}[t]{0.49\linewidth}
    \vspace{0pt}
    \captionsetup{justification=raggedright,singlelinecheck=false}
    \captionof{table}{Reported training compute of video alignment methods.}
    \label{tab:compute}
    \vspace{2pt}
    \scriptsize
    \setlength{\tabcolsep}{2pt}
    \begin{tabular*}{\linewidth}{@{\extracolsep{\fill}}llcc@{}}
        \toprule
        Method & Model & GPUs & Cost \\
        \midrule
        DanceGRPO & HunyuanVideo & 16--32 H800 & --- \\
        PRFL      & Wan2.1-14B   & ---         & 51.1\,s/step \\
        \method   & Wan2.1-1.3B  & 4 H100      & 2.1\,h \\
        \bottomrule
    \end{tabular*}

    \vspace{2pt}
    {\scriptsize DanceGRPO~\citep{xue2025dancegrpo};
    PRFL~\citep{mi2026video}.}
\end{minipage}
\end{table}

\section{Conclusion}
We introduced \method, which co-evolves the reward model with the generator to mitigate this failure and stabilize optimization. This work has the following limitations: (i) we conduct experiments on 1.3B-scale model; training larger models and developing finer-grained probes of latent reward exploitation may reveal additional insights. (ii) We use a limited set of datasets, and future work should test \method on more challenging benchmarks.

\subsection*{AI use statement}
In this work, we used generative AI tools for language polishing, LaTeX formatting, and literature search.
We have reviewed all AI-assisted work and take responsibility for the final content of this work,
including text or artifacts produced with the aid of generative AI.

\subsection*{Reproducibility statement}
The training setup and hyperparameters are described in Sec.~\ref{sec:exp}; the fixed-reward
baseline and additional analyses are given in the appendix. Code will be released as
supplementary material.

\bibliography{iclr2027_conference}
\bibliographystyle{iclr2027_conference}

\newpage
\appendix
\section{OOD Diagnostics in the Reward-Feature Space}
\label{app:ood_latent}

\begin{figure}[h]
    \centering
    \begin{minipage}[t]{0.48\linewidth}
        \centering
        \includegraphics[width=\linewidth]{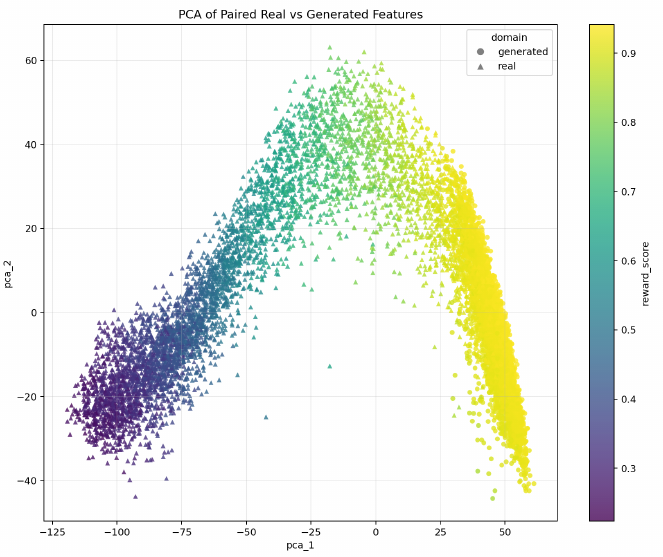}\\[-2pt]
        {\small (a) PCA}
    \end{minipage}\hfill
    \begin{minipage}[t]{0.48\linewidth}
        \centering
        \includegraphics[width=\linewidth]{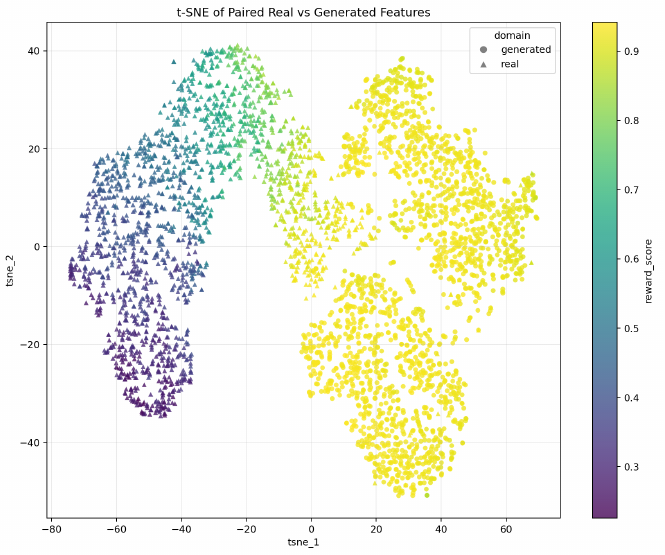}\\[-2pt]
        {\small (b) t-SNE}
    \end{minipage}
    \caption{\textbf{Real vs.\ generated samples in the reward-feature space under a fixed reward model}, colored by reward score ($\blacktriangle$ real, $\bullet$ generated). The two domains are almost perfectly separated, and generated samples concentrate in a high-reward region (yellow) beyond the range covered by real samples.}
    \label{fig:ood_latent}
\end{figure}

Fig.~\ref{fig:ood_latent} repeats the analysis of Fig.~\ref{fig:ood_tsne} on a larger set of samples from a fixed-reward run, visualizing the features fed to the reward head. The two domains are almost perfectly separable, and only $1.3\%$ of each sample's nearest neighbors come from the other domain, so the separation is local as well as global. Real samples span the full range of reward scores along an ordered direction, whereas generated samples extend beyond its high-score end and receive uniformly high rewards, in a region where the reward model received no supervision.

\begin{figure}[t]
\centering
\begin{minipage}[t]{0.32\linewidth}
  \centering
  \includegraphics[width=\linewidth]{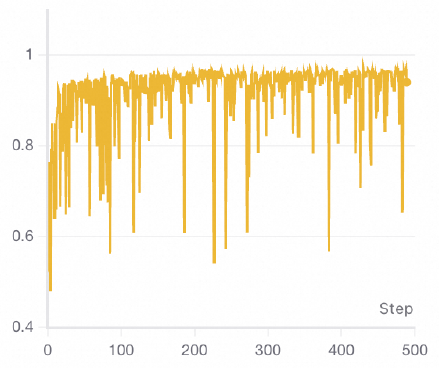}
  \phantomsubcaption\label{fig:reward_curves_a}\\[-2pt]
  {\small (a) w/o flow-matching anchor}
\end{minipage}\hspace{0.03\linewidth}
\begin{minipage}[t]{0.32\linewidth}
  \centering
  \includegraphics[width=\linewidth]{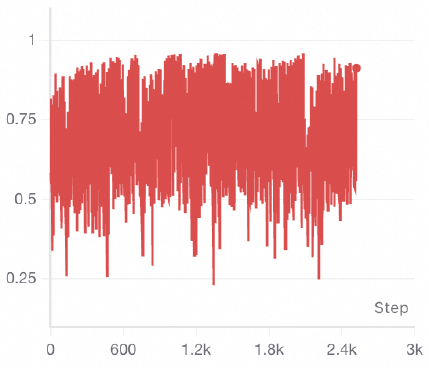}
  \phantomsubcaption\label{fig:reward_curves_b}\\[-2pt]
  {\small (b) w/ flow-matching anchor}
\end{minipage}
\vspace{-4pt}
\caption{\textbf{Training reward against a fixed latent reward model.}
(a) Without regularization, the reward saturates near 0.95 within about 20 updates,
while the decoded videos blur by step 15, show grid-like artifacts by step 24, and
collapse by step 100.
(b) With the flow-matching anchor, the reward no longer saturates but also shows no
clear upward trend over 2.5k updates. A fixed reward thus leaves two outcomes:
rapid exploitation, or little effective optimization.}
\label{fig:reward_curves}
\end{figure}

\section{Layer-wise Reward Features}
\label{app:layer_analysis}

Tab.~\ref{tab:layer_analysis} compares reward models built on features from different DiT blocks. Block 8 gives the highest held-out preference accuracy.

\begin{table}[h]
\centering
\caption{\textbf{Reward models built on different DiT blocks.} Preference acc.: held-out accuracy on VideoDPO pairs. Reward--quality corr.: Pearson correlation between reward scores and VBench imaging quality on generated videos. Stable steps: number of generator updates before dynamic degree falls below $0.3$.}
\label{tab:layer_analysis}
\small
\setlength{\tabcolsep}{4pt}
\begin{tabular}{cccc}
\toprule
DiT block & Preference acc. $\uparrow$
& Reward--quality corr. $\uparrow$
& Stable steps $\uparrow$ \\
\midrule
4  & 68.2 & 0.41  & 420 \\
8  & 76.4 & 0.54 & 850 \\
12 & 74.1 & 0.43 & 510 \\
16 & 71.6 & 0.37 & 290 \\
\bottomrule
\end{tabular}
\end{table}

\section{Implementation Details}
\label{app:impl}
The generator is sampled on a 40-step schedule. We use AdamW~\citep{loshchilov2017decoupled} with constant learning rates of $2\times10^{-6}$ (generator) and $3\times10^{-6}$ (reward head), a flow-matching weight $\lambda_{\mathrm{fm}}{=}1.5$, and evaluate the EMA of the generator (decay $0.99$). All co-evolving variants share the same warm start: starting from the pretrained model, the reward head is trained for 200 steps with the generator frozen, the generator is then trained for 100 steps with the absolute-target objective, and the reward head is re-fit for another 100 steps. From this common checkpoint, each variant trains the generator for 100 further steps with the objective under study.

\section{All Ablation Study}
\label{sec:all_ablation}
\begin{table}[t]
\centering
\caption{VBench scores for all runs.
SC = subject\_consistency, BC = background\_consistency, MS = motion\_smoothness,
DD = dynamic\_degree, AQ = aesthetic\_quality, IQ = imaging\_quality.}

\label{tab:vbench_all}
\small
\setlength{\tabcolsep}{4.5pt}
\begin{tabular}{llcccccc}
\toprule
Experiment & Ckpt & SC & BC & MS & DD & AQ & IQ \\
\midrule
Fixed LRM & 500 & 0.969 & 0.972 & 0.983 & 0.681 & 0.642 & 0.600 \\
\midrule
v2         & 300  & 0.956 & 0.961 & 0.969 & 0.896 & 0.596 & 0.638 \\
v2         & 400         & 0.966 & 0.967 & 0.975 & 0.778 & 0.619 & 0.656 \\
step200    & 200         & 0.964 & 0.970 & 0.987 & 0.681 & 0.603 & 0.679 \\
v2         & 750         & 0.978 & 0.982 & 0.992 & 0.194 & 0.631 & 0.629 \\
\midrule
w1p5       & 350         & 0.946 & 0.960 & 0.984 & 0.837 & 0.567 & 0.627 \\
w1p5       & 400         & 0.963 & 0.969 & 0.985 & 0.708 & 0.608 & 0.628 \\
w1p5       & 450         & 0.971 & 0.976 & 0.988 & 0.167 & 0.636 & 0.576 \\
\midrule
\multicolumn{8}{l}{\textit{dyn300 (new dynamic dataset)}} \\
dyn300     & 325         & 0.960 & 0.969 & 0.982 & 0.750 & 0.587 & 0.647 \\
dyn300     & 350         & 0.960 & 0.966 & 0.981 & 0.694 & 0.603 & 0.639 \\
dyn300     & 375         & 0.954 & 0.962 & 0.981 & 0.667 & 0.594 & 0.634 \\
dyn300     & 400         & 0.967 & 0.974 & 0.988 & 0.472 & 0.601 & 0.638 \\
\midrule
\multicolumn{8}{l}{\textit{dyn300 + EMA}} \\
dyn300ema  & 325         & 0.960 & 0.968 & 0.982 & 0.778 & 0.587 & 0.637 \\
dyn300ema  & 350         & 0.959 & 0.967 & 0.980 & 0.750 & 0.594 & 0.644 \\
dyn300ema  & 375         & 0.959 & 0.966 & 0.983 & 0.667 & 0.598 & 0.631 \\
dyn300ema  & 400         & 0.966 & 0.968 & 0.985 & 0.583 & 0.604 & 0.620 \\
\midrule
\multicolumn{8}{l}{\textit{oldw400ema (re-warmup 400 on VideoDPO, EMA)}} \\
oldw400ema & 450         & 0.971 & 0.974 & 0.982 & 0.611 & 0.611 & 0.661 \\
oldw400ema & 500         & 0.967 & 0.973 & 0.983 & 0.569 & 0.612 & 0.641 \\
\midrule
\multicolumn{8}{l}{\textit{relgap (same warmed ckpt-400, relative real--fake gap G loss, EMA)}} \\
relgapema  & 450         & 0.963 & 0.968 & 0.981 & 0.722 & 0.605 & 0.650 \\
relgapema  & 500         & 0.959 & 0.966 & 0.977 & 0.694 & 0.592 & 0.659 \\
\midrule
\multicolumn{8}{l}{\textit{relgap + kernel anchor (same warmed ckpt-400, EMA)}} \\
kernelema, $\lambda_k{=}0.05$ & 600 & 0.914 & 0.948 & 0.975 & 0.819 & 0.594 & 0.612 \\
kernelema, $\lambda_k{=}0.03$ & 600 & 0.912 & 0.947 & 0.975 & 0.847 & 0.587 & 0.615 \\
\bottomrule
\end{tabular}
\end{table}
Tab.~\ref{tab:vbench_all} collects the six VBench quality dimensions for
every run: subject consistency (SC), background consistency (BC), motion
smoothness (MS), dynamic degree (DD), aesthetic quality (AQ), and imaging
quality (IQ), each benchmark on VBench. The runs are as
follows. \emph{Fixed LRM} is the pipeline with a frozen latent reward model~\citep{mi2026video}. \emph{v2} and \emph{w1p5} are co-evolving runs started from
the pretrained model with an absolute score target, differing only in the
generator learning rate ($2{\times}10^{-6}$ and $5{\times}10^{-6}$); the
first 200 steps of v2 update the reward head alone, so its step-200
checkpoint is the pretrained model. \emph{dyn300} resumes v2 at step 300
on VisionReward motion pairs instead of VideoDPO, with and without an
EMA generator. \emph{oldw400ema}, \emph{relgapema} and \emph{kernelema}
share one warm start: v2 at step 300 with the reward head re-fit for a
further 100 steps. They then train the generator for 100 steps with the
absolute target, the relative gap, and the relative gap plus kernel anchor
respectively, and are evaluated with the EMA generator.

Two baselines bracket the results. The pretrained model sets the reference
for IQ (0.679); the fixed reward leaves DD
unchanged and lowers IQ by eight points. The naive co-evolving runs first
raise DD (0.896 for v2 at step 300) and then collapse it (0.194 at step
750; 0.167 for w1p5 at step 450). The rise in SC, BC and MS along the way
reflects videos that stop moving, not an improvement. Among the runs with
the shared warm start, the absolute target continues the decline in DD
(0.569). The relative gap holds DD at the pretrained level (0.694) with the
best IQ of the block (0.659).
\begin{table}[H]
    \centering
    \small
    \setlength{\tabcolsep}{4pt}
    \caption{Generator objective, evaluated at two checkpoints of the same 100-step run.}
    \label{tab:abl}
    \begin{tabular}{lcccccc}
        \toprule
        Objective & Step & Dyn. & Imaging & Aesth. & Subj. & Smooth \\
        \midrule
        Absolute target & 450 & 61.11 & 66.13 & 61.05 & 97.14 & 98.19 \\
                        & 500 & 56.94 & 64.09 & 61.24 & 96.74 & 98.28 \\
        Relative gap    & 450 & 72.22 & 64.96 & 60.47 & 96.25 & 98.14 \\
                        & 500 & 69.44 & 65.85 & 59.23 & 95.88 & 97.69 \\
        \bottomrule
    \end{tabular}
\end{table}

\end{document}